\def\year{2019}\relax
\documentclass[letterpaper]{article} %DO NOT CHANGE THIS
\usepackage{aaai19}  %Required
\usepackage{times}  %Required
\usepackage{helvet}  %Required
\usepackage{courier}  %Required
\usepackage{url}  %Required
\usepackage{graphicx}  %Required
\usepackage{amsmath}
\usepackage{amsfonts}
\usepackage{subcaption}

\begin{document}
% The file aaai.sty is the style file for AAAI Press proceedings, working notes, and technical reports.

\title{Class-specific Anchoring Proposal for 3D Object Recognition in LIDAR and RGB Images}
\author{Amir Hossein Raffiee\\
{School of Mechanical Engineering, Purdue University}\\
araffie@purdue.edu
\And
Humayun Irshad\\
{Figure Eight Technology Inc}\\
humayun.irshad@figure-eight.com}
\maketitle

\begin{abstract}
Detecting objects in a two-dimensional setting is often insufficient in the context of real-life applications where the surrounding environment needs to be accurately recognized and oriented in three-dimension (3D), such as in the case of autonomous driving vehicles. Therefore, accurately and efficiently detecting objects in the three-dimensional setting is becoming increasingly relevant to a wide range of industrial applications, and thus is progressively attracting the attention of researchers. Building systems to detect objects in 3D is a challenging task though, because it relies on the multi-modal fusion of data derived from different sources. In this paper, we study the effects of anchoring using the current state-of-the-art 3D object detector and propose Class-specific Anchoring Proposal (CAP) strategy based on object sizes and aspect ratios based clustering of anchors. The proposed anchoring strategy significantly increased detection accuracy's by 7.19\%, 8.13\% and 8.8\% on Easy, Moderate and Hard setting of the pedestrian class, $2.19\%$, $2.17\%$ and $1.27\%$ on Easy, Moderate and Hard setting of the car class and 12.1\% on Easy setting of cyclist class. We also show that the clustering in anchoring process also enhances the performance of the regional proposal network in proposing regions of interests significantly. Finally, we propose the best cluster numbers for each class of objects in KITTI dataset that improves the performance of detection model significantly.     
\end{abstract}

%===========================================================
\section{Introduction}
%===========================================================

Nowadays, the object detection and recognition task is carried out widely using convolutional neural networks (CNN). In this regard, many models have been developed to detect 2D objects in RGB images based on the extraction of features of the image using CNNs \cite{girshick2015fast,girshick2014rich,liu2016ssd,redmon2016you}. While these models are capable of achieving significant performance in object detection and recognition in many areas, but these models are unable to show reasonable performance in few areas, particularly in object detection in 3D autonomous driving. 

For instance, in order to allow safe motion planning for self-driving cars, the surrounding environment must be recognized in a 3D manner; additionally, information about the location, size and orientation of the surrounding objects should also be provided. For the purpose of the recognition, various types of information can be used such as depth information and distance information from other sources like 3D point cloud. The work presented in \cite{kehl2017ssd} used RGB image to localize and estimate the orientation of the objects. Furthermore, RGB-D images can also be used, allowing the models to fuse the depth information with a RGB image \cite{song2016deep,song2014sliding} leading to a substantial performance gain compared to the models that solely use RGB images. However, with the recent improvements in LIDAR (Light Detection And Ranging) technology, predictive models have been developed to estimate the 3D bounding boxes using point clouds produced by LIDAR sensors \cite{qi2017frustum,chen2017multi,zhou2017voxelnet,yang2018pixor,simon2018complex}. 

Because the point cloud is an unstructured collection of points in a 3D space, it is difficult to apply standard convolutional operations. Hence, different approaches have been proposed to deal with this problem. In this regard, the point cloud can be represented in a 3D voxel grid space where the value of each cell is either a scalar or a vector variable representing the hand-crafted features of the points falling in that grid. In this approach, 3D convolutions can be applied to extract the high-level features of the point cloud that lead to high computation due to the high dimensionality of the problem. Furthermore, the point cloud can be projected into multiple planes \cite{engelcke2017vote3deep,wang2015voting,li2016vehicle}. 

Another approach is the projection of the point cloud onto a 2D plane such as birds-eye-view plane and then its discretization into equally spaced grid cells. The value of the grid cells is determined by using hand-crafted features such as density, maximum height and intensity. This way, the 3D point cloud is transformed into a 2D image-based data on which 2D convolution can be applied for feature extraction. Two problems associated with this method are (i), the loss of information during the projection of points from 3D to 2D and ii), the need to manually craft features to represent them in an image-based fashion \cite{chen2017multi,ku2017joint,yang2018pixor}. 

\cite{chen2017multi} proposes to fuse the information from the 2D-projected point cloud on birds-eye-view and front map with RGB camera data to compensate for that loss of information. In order to avoid the need for hand-crafted features in the representation of the point cloud, the authors of \cite{qi2017pointnet,qi2017pointnet++} developed networks that treat the point cloud in 3D space directly without the need to transform them into another space. 

In \cite{ku2017joint}, the authors present the aggregate view object detection (AVOD) model powering the current state-of-the-art 3D object detector on the KITTI data set, the 3D object detection benchmark \cite{geiger2013vision}. In the current paper, we set to use the AVOD framework in order to evaluate the performance improvement in 3D object detection using class-specific anchoring proposals. This framework is composed of three consecutive networks: (i) encoder-decoder network which extracts feature maps for LIDAR points and RGB images, (ii) 3D Region Proposal Network (RPN) that includes a region-based fusion network which combined RGB and BEV feature maps to compensate for the information loss occurring during the 2D transformation phase of the point cloud and select a list of top proposed regions and (iii) final detection and pose estimation network that regresses the modification and orientation of selected regions and categorizes the classification label for selected regions. In our proposed study, we are extending AVOD framework by adding class-specific anchoring proposal to improve the anchoring strategy by addressing the issue of classes which have objects with large variations in sizes and aspect ratios. 

Due to the variety of sizes and aspect ratios of the objects in the class of cyclist and pedestrian in KITTI data set, the current object detectors have poor performance in proposing appropriate anchor size that leads to poor performance in the final object detection. As a remedy, inspired by 2D object detectors we used $K-$mean and Gaussian Mixture Model  clustering methods to have better prior information about the object sizes in which we can set the number clusters. This capability can be used for the class of objects that has large variance in their size and aspect ratio which helps the framework to gain performance in detection for those objects.  

%===========================================================
\section{Related Works}
%===========================================================

The F-PointNet \cite{qi2017frustum} achieves notable performance for 3D object detection and birds-eye-view detection on cars, pedestrians and cyclists on the KITTI benchmark suite. This method uses a 2D Faster RCNN object detector to find 2D boxes including the object on RGB camera image. Subsequently, the detected boxes are extruded to identify the point cloud falling into the frustum corresponding to the boxes. The discovered point cloud is classified in a binary fashion to separate the points constructing the object of interest and 3D regression is conducted on the separated points. The main drawback of this method comes from the fact that the accuracy of the model is highly dependent on the accuracy of the 2D object detector on RGB image. For instance, if the 2D detector misses the object on RGB image, the second Network is not able to localize the missed object in the 3D space. Furthermore, the consecutive nature of this model (2D detector Network followed by 3D detector) extends the inference time, which is a noteworthy issue in the context of numerous applications, including autonomous vehicles.

Multi-view 3D \cite{chen2017multi} (MV3D) is a 3D object detector that fuses the information from three separate sources: the projected point cloud on birds-eye-view, the front-view map and the RGB camera image. In this method, the LIDAR points are mapped in voxel grids, and handcrafted features such as maximum height and density are used to convert unordered point clouds to the voxel grid map. The input feature maps are fed into the 3D Region Proposal Network that uses the combined information from these three sources in order to achieve higher recall compared to the cases where only one of the sources is used. The main challenge with this method is the high cost of computation which also increases the inference time.

%===========================================================
\section{Methodology}
\begin{figure*}[h]
    \centering
    \includegraphics[height=2.8in,width=6.6in]{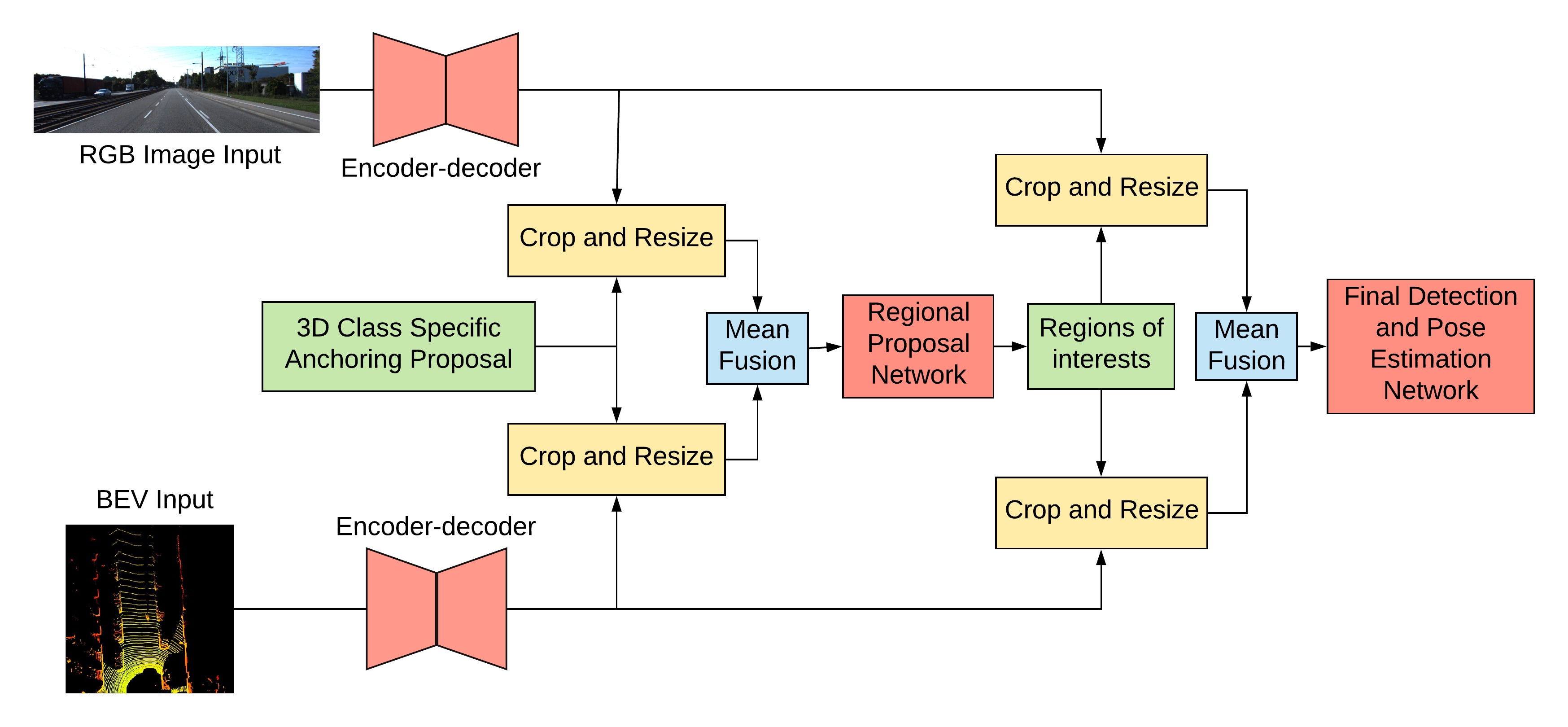}
    \caption{Architecture of Class-specific Anchoring Proposal (CAP)-AVOD used for 3D objects recognition.}
    \label{Network} 
\end{figure*}

%===========================================================
In this study, we explore the effect of Class-specific Anchoring Proposal on AVOD algorithm \cite{ku2017joint} for 3D object recognition with the use of RGB images and point cloud data illustrated in Figure \ref{Network}. The framework takes two inputs (RBG image and corresponding 3D point cloud) and predicts 3-dimensional bounding boxes and classifies the objects into three classes of cyclist, pedestrian and car. The main components of the framework are Bird-Eye-View (BEV) map generation, class-specific anchoring proposal (CAP) strategy and AVOD model which contains three networks: encoder-decoder network, Region Proposal Network (RPN) and final detection and pose estimation network. 

\subsection{BEV Map Generation}

Following \cite{chen2017multi}, 2D representation of LIDAR data points is used to generate the BEV map. In this process, the 3D data points are projected onto a series of planes parallel to the ground plane. Each plane is discretized with the resolution of $0.1m$ in $X$ and $Z$ directions ($X$ and $Z$ aligns with the camera 0 coordinates \cite{geiger2013vision} in KITTI dataset) to create 2D voxel grids. The voxel grid value is encoded by the maximum height of the LIDAR point trapped in that grid. The resulting tensor represents the height map of the point cloud in 3D space scene. Furthermore, the density of the points in each grid is computed and concatenated to the height map of LIDAR points to construct the BEV map.  Hence, the input tensor fed to the model is of the shape of ($H,W,C+1$) where $H$ and $W$ represents the height and the width of the 2D voxel grid and $C$ is the number of planes that data point are projected \cite{chen2017multi,yang2018pixor} plus one channel that contains point density information computed per cell as $min(1.0, \frac{\log(N+1)} {\log 16})$. It should be noted that the points are cropped at [-40,40] $\times$ [0,80] meter in $X$ and $Z$ directions, respectively, which encloses larger area compared to previous works \cite{chen2017multi,ku2017joint}. We noticed that significant number of objects (particularly objects belonging to the cyclist and pedestrian classes) falls outside the area considered in their model which affects the the model performance negatively and increases the number of missed objects.

\subsection{Class-specific Anchoring Proposal (CAP)}

The AVOD model show acceptable performance on detecting cars \cite{ku2017joint}, while its performance drops significantly when there is a large variations in size and aspect ratios between objects such as cyclists and pedestrians. In this work, we are exploring the effect of anchoring on the performance of the model for detecting objects in three classes of cyclist, pedestrian and car.

For AVOD model, we study the effect of proposed anchoring on the performance of the two networks; RPN which selects anchors and produces Regions of Interest (ROI) and final detection and pose estimation network which recognizes and localizes objects in proposed ROIs. In multi-stage object detectors, the anchors' size are considered as prior information about the size of the desired objects to be detected by the model. Hence, having an appropriate prior information about the size of the objects improves the performance of the model significantly. In this strategy, class specific anchors are generated based on each class prior information. The number of the anchors proposed per frame is approximately $80$ to $100$K for AVOD and CAP-AVOD. 

In order to find appropriate sizes for class specific anchors, we use $K$-mean clustering and Gaussian Mixture Model (GMM) methods. As we know, the size of the objects in 3D can be represented by length ($L$), width ($W$) and height ($H$). Therefore, each object in particular class is considered as a vector $\textbf{x}$ with three features ($L,H,W$). In this method, the objects in each class are clustered into specific number of groups ($n$) and the mean values of each group is considered as the size of the proposed anchors. Each object in class $C$ is represented by $x_{ic}=(L_{ic},H_{ic},W_{ic})$. Hence, given $m$ objects for class $C$ in training data we can represent all the objects in this class as:

\begin{equation}\label{eq:0-1}
    X_{c}=(x_{1c},x_{2c},...,x_{mc})
\end{equation}

Using $K$-mean clustering method, we try to find clusters $S_{jc}$ ($j\in[1,n]$) for each class of objects ($C$) by minimizing the variance in each cluster as follow:

\begin{equation}\label{eq:0-2}
    \underset{S_{jc}}{argmin}\sum_{j=1}^{n}\sum_{x_{ic}\in S_{jc}}{||x_{ic}-\mu_{jc}||}^2
\end{equation}
where $\mu_{jc}$ is the mean of data points belonging into $S_{jc}$ cluster. 

In the second method, Expectation Maximization (EM) techniques is exploited for GMM to find the best clusters in the training data. In this technique, $\pi_{jc}$, $\mu_{jc}$ and $\sum_{jc}$ that denote weight, mean vector and covariance matrix of a given j-th cluster for class $C$, respectively, are updated iterativly in the following steps:
\begin{equation}\label{eq:0-2}
    \gamma_{jc}(x)=\frac{\pi_{jc} N(x|\mu_{jc},\sum_{jc})}{\sum_{i=1}^{n}{\pi_{ic}}N(x|\mu_{ic},\sum_{ic})}
\end{equation}
\begin{equation}\label{eq:0-2}
    \mu_{jc}=\frac{\sum_{i=1}^{N_c}{\gamma_{jc}(x_i)x_i}}{\sum_{i=1}^{N_c}{\gamma_{jc}(x_i)}}
\end{equation}
\begin{equation}\label{eq:0-2}
    \sum_{jc}=\frac{\sum_{i=1}^{N_c}{\gamma_{jc}(x_i)(x_i-\mu_{jc})(x_i-\mu_{jc})^T}}{\sum_{i=1}^{N_c}{\gamma_{jc}(x_i)}}
\end{equation}
\begin{equation}\label{eq:0-2}
    \pi_{jc}=\frac{\sum_{i=1}^{N_c}{\gamma_{jc}}}{N_c}
\end{equation}
where $N_c$ shows the number of objects in class $C$.

\subsection{CAP-AVOD framework}

 The CAP-AVOD framework contains three networks; (i) encoder-decoder network, (ii) Region Proposal Network (RPN) and (iii) final detection and pose estimation network. We trained two encoder-decoder networks \cite{lin2017feature} which output two feature maps, one for RGB image and one for BEV map. The use of bottom-up decoder for feature extraction enables the generated feature map to retain the global and local information of the input tensor. Similar to multi-stage 2D object detectors, the RPN is also fed by proposed anchors. The proposed anchors are cropped and resized from the RGB and BEV feature maps and the resulted tensors are fused to maintain the information from two sources. Based on the proposed anchors, RPN generates Region Of Interests (ROIs). The selected ROIs cover the regions which contains the desired objects to be detected. The final detection and pose estimation network regresses the offset of selected ROIs and determines the class categories of each ROI. The architecture of CAP-AVOD is illustrated in Figure \ref{Network}.

\subsection{Model evaluation}

The evaluation of the model is conducted in three aspects. First, we evaluate the anchoring process and compute how well the proposed anchors cover the objects of interest. Second, the performance of RPN is evaluated by computing the recall number based on the selected ROIs and ground truth objects. Third, the overall performance of the model is studied according to 3D Average Precision ($AP$). As it is noted, the proposed anchors fed into the model is modified in the RPN to cover the objects in the 3D space. This process indicates the significance of anchor proposal method for object detection task. If the proposed anchors cover a good portion of objects in the 3D space, the RPN modifies the anchors efficiently with small amount of offset and the probability of missing objects reduces remarkably. 

\begin{figure}[h]
    \centering
    \includegraphics[height=1.3in,width=3.3in]{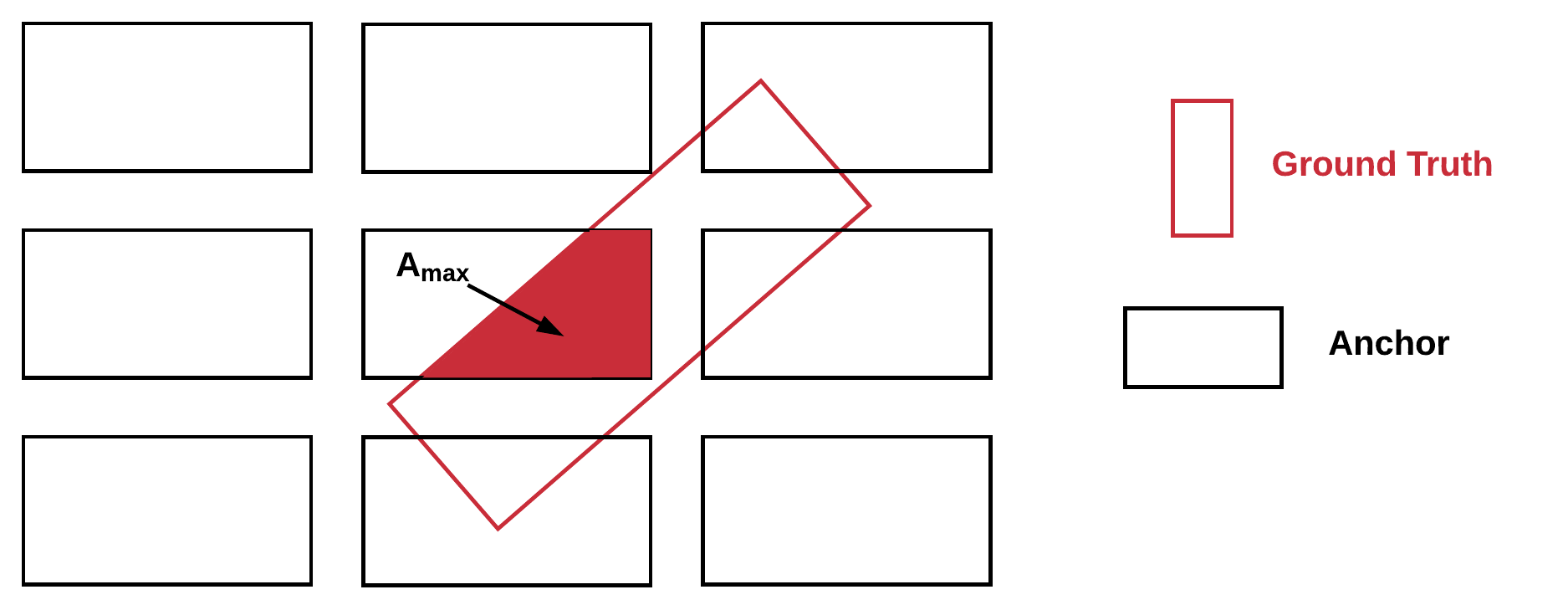}
    \caption{Overlapped area of ground truth box and proposed anchor.}
    \label{Achoring_evaluation}
\end{figure}

In order to evaluate the anchoring of the 3D space, the fraction of ground truth object overlapped by the anchors is computed. According to Figure \ref{Achoring_evaluation}, the overlapped area Equation .\ref{eq:1} is calculated by division of maximum overlapped area of the ground truth object ($A_{max}$) by the entire area of the ground truth object ($A_{GT}$). The area of the objects and anchors are computed in 2D as they are considered in BEV.       

\begin{equation}\label{eq:1}
    Overlapped\ Area = \frac{A_{max}}{A_{GT}}
\end{equation}

%===========================================================
\section{Results and Discussion}
%===========================================================

The KITTI training data set is split into a training set and a validation set (using roughly a 50-50 split) following the process mentioned in \cite{xu2017pointfusion,chen2017multi}. All the evaluation of proposed CAP and AVOD framework is performed on validation set. 

\subsection{Anchoring evaluation}\label{anchoringevaluation}

The anchoring process in multi-stage object detectors proposes the potential locations of the objects that is tuned by the RPN to focus the attention of the last network to those regions. Therefore, the model does not need to explore the entire area to detect the objects which significantly reduces the computational cost. Hence, proposing anchors that covers descent portion of the objects enables the RPN and final detection and pose estimation network to predict and classify objects more accurately and regress the ROIs more efficiently \cite{ren2015faster,he2017mask}. Accordingly, for the proposed anchors fed to RPN that does not significantly cover the object of interest the model can not detect the objects in the later stages. This effect shows that the multi-stage object detector requires an efficient and smart anchoring process which reduces the computational cost allowing real-time inference and also reduces the probability of missing objects leading to the improvement in recall of the 3D Region Proposal Network. 

\begin{figure}
    \centering
    \begin{subfigure}{.5\textwidth}
        \centering
        \includegraphics[height=1.45in,width=1.45in]{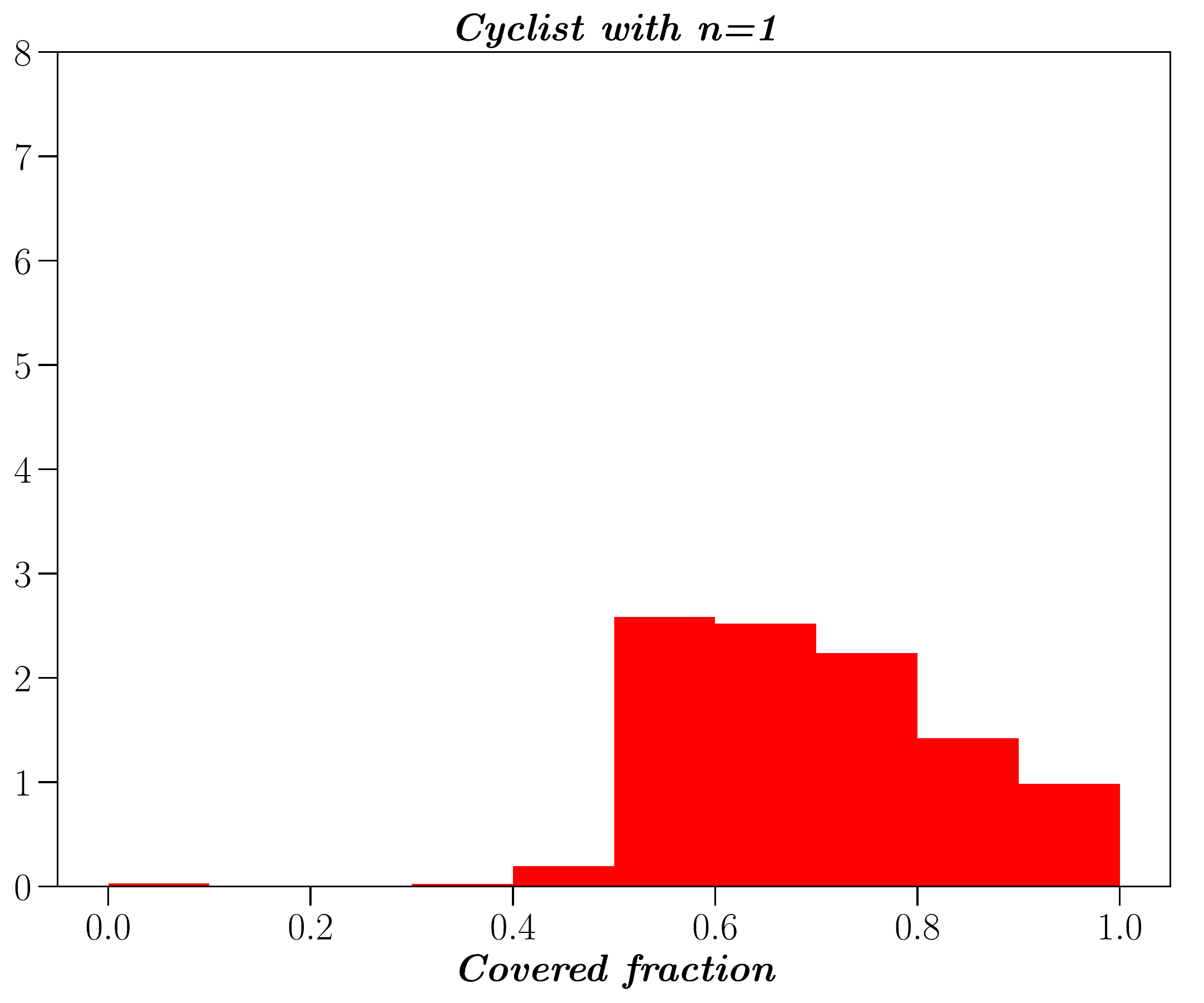}
        \includegraphics[height=1.45in,width=1.45in]{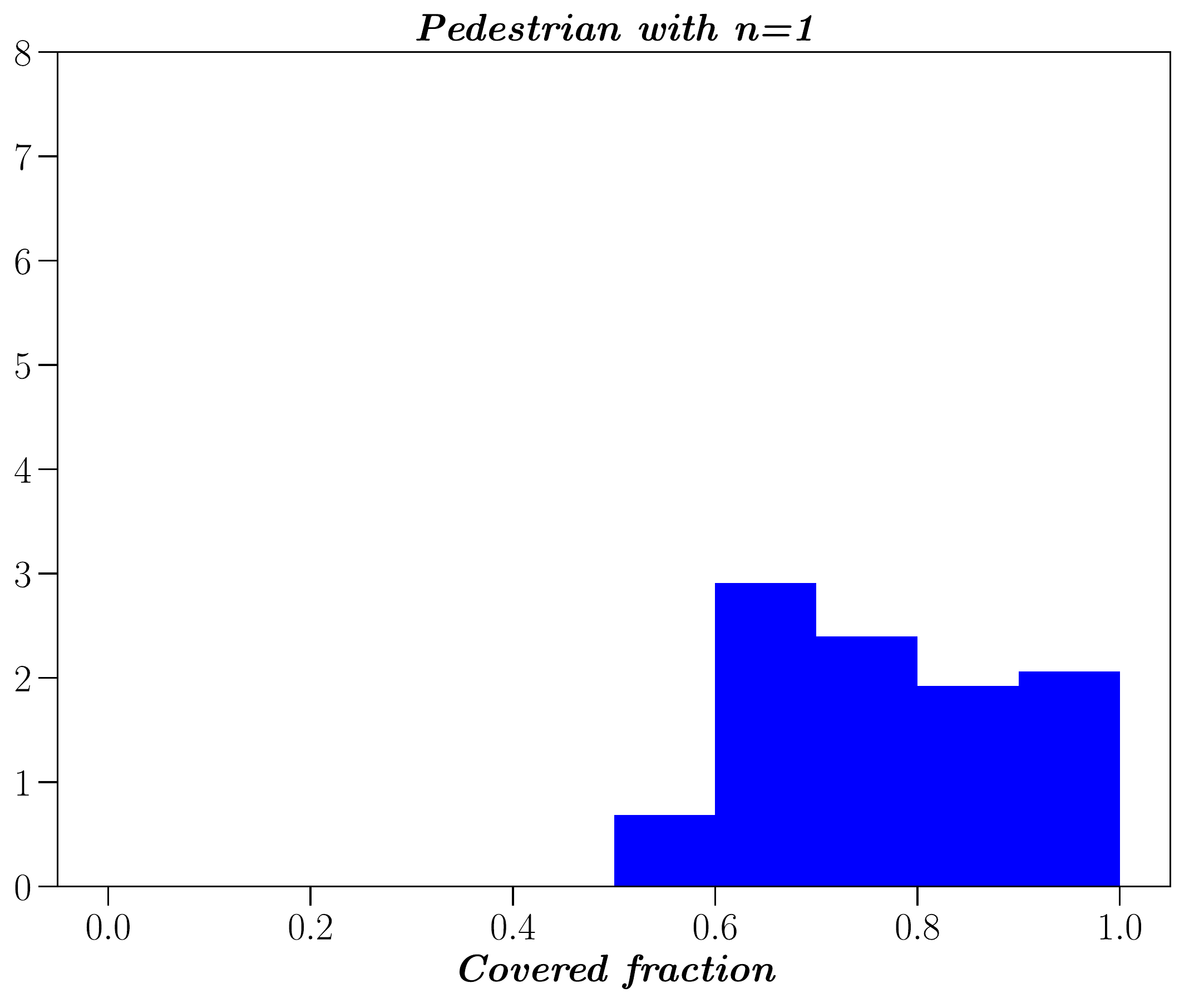}
        \caption{1 Cluster of K-Mean Clustering (n=1)}
    \end{subfigure}
    \begin{subfigure}[t]{.5\textwidth}
        \centering
        \includegraphics[height=1.45in,width=1.45in]{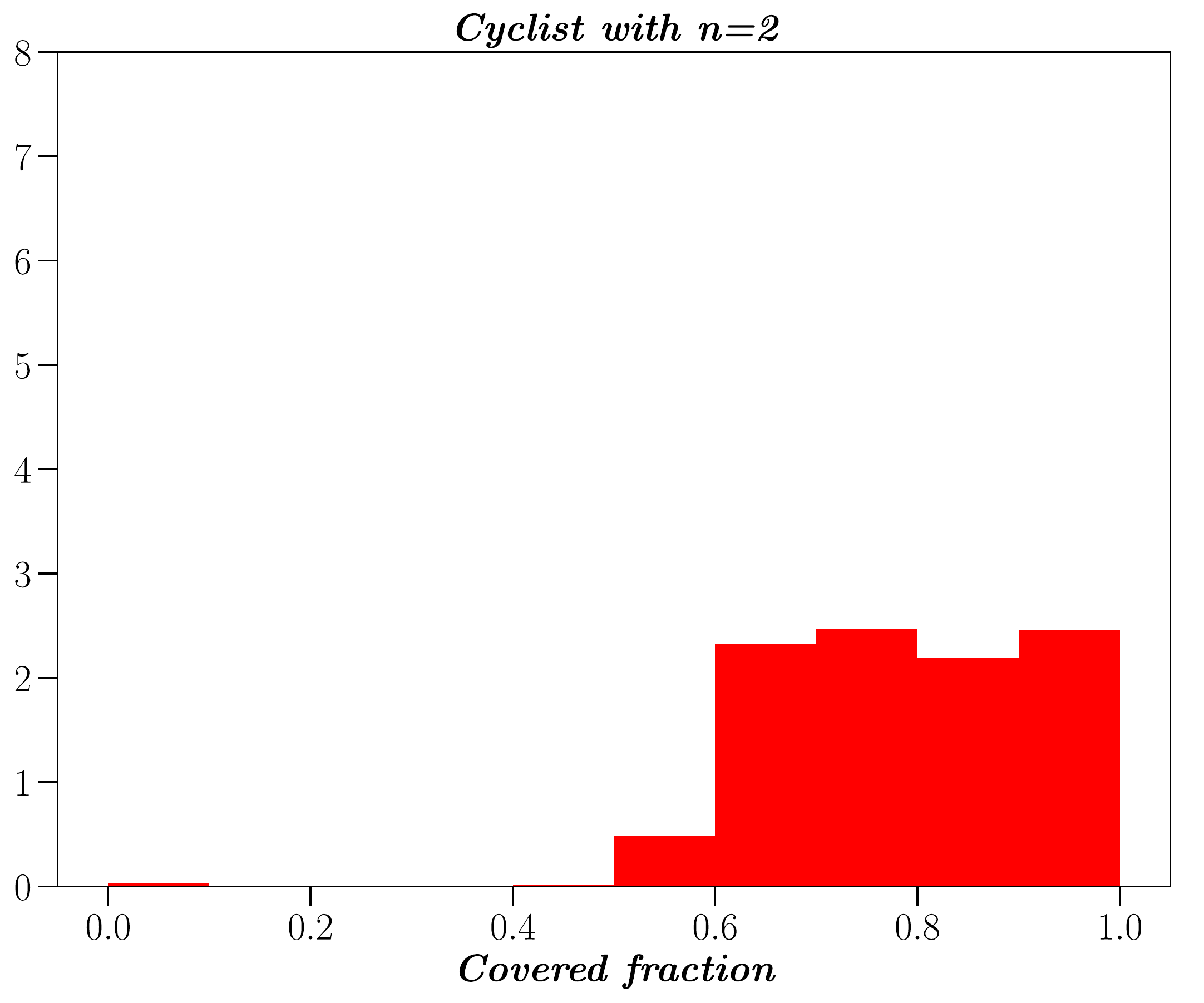}
        \includegraphics[height=1.45in,width=1.45in]{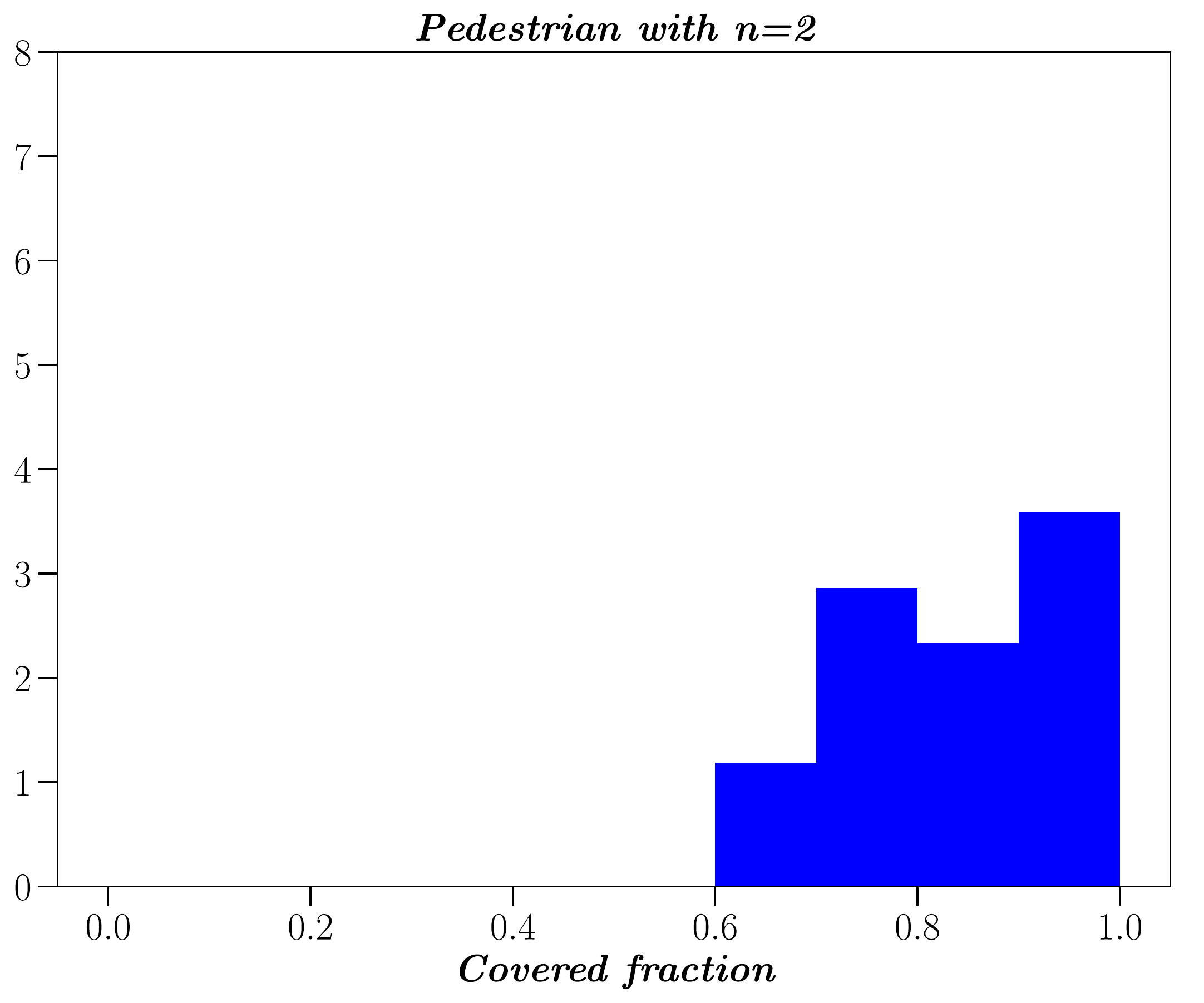}
        \caption{2 Clusters of K-Mean Clustering (n=2)}
    \end{subfigure}
    \begin{subfigure}[t]{.5\textwidth}
        \centering
        \includegraphics[height=1.45in,width=1.45in]{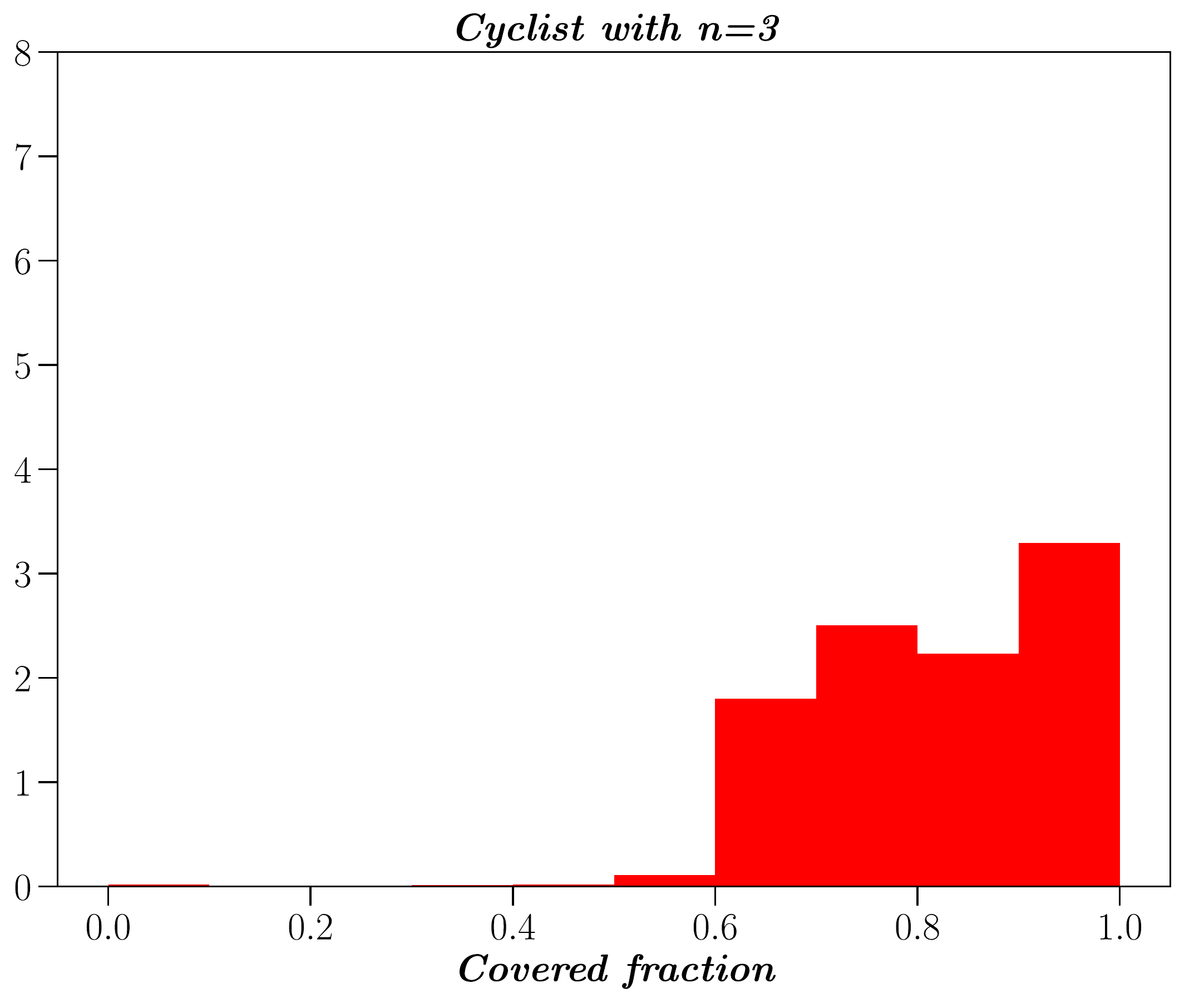}
        \includegraphics[height=1.45in,width=1.45in]{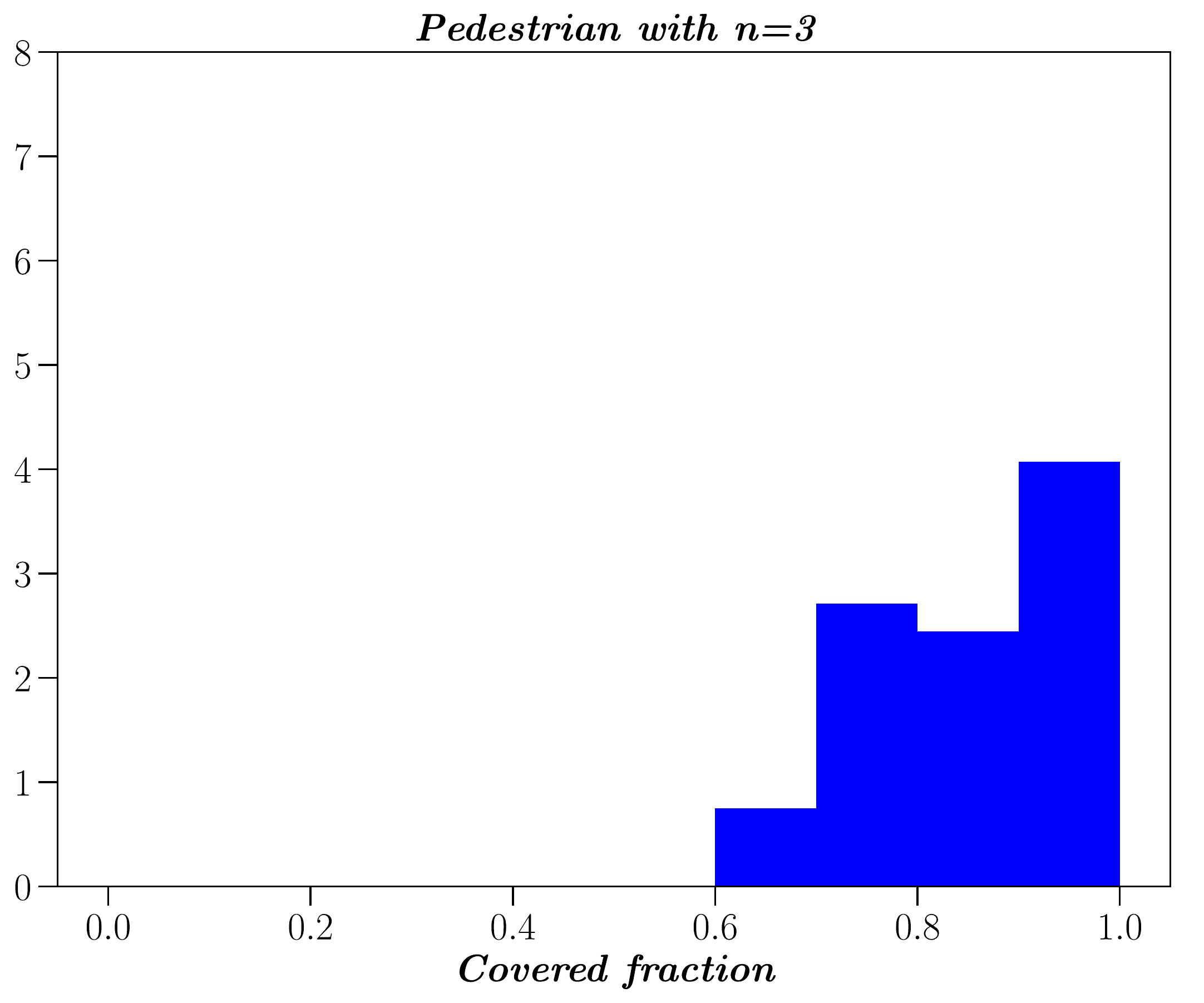}
        \caption{3 Clusters of K-Mean Clustering (n=3)}
    \end{subfigure}
    \begin{subfigure}{.5\textwidth}
        \centering
        \includegraphics[height=1.45in,width=1.45in]{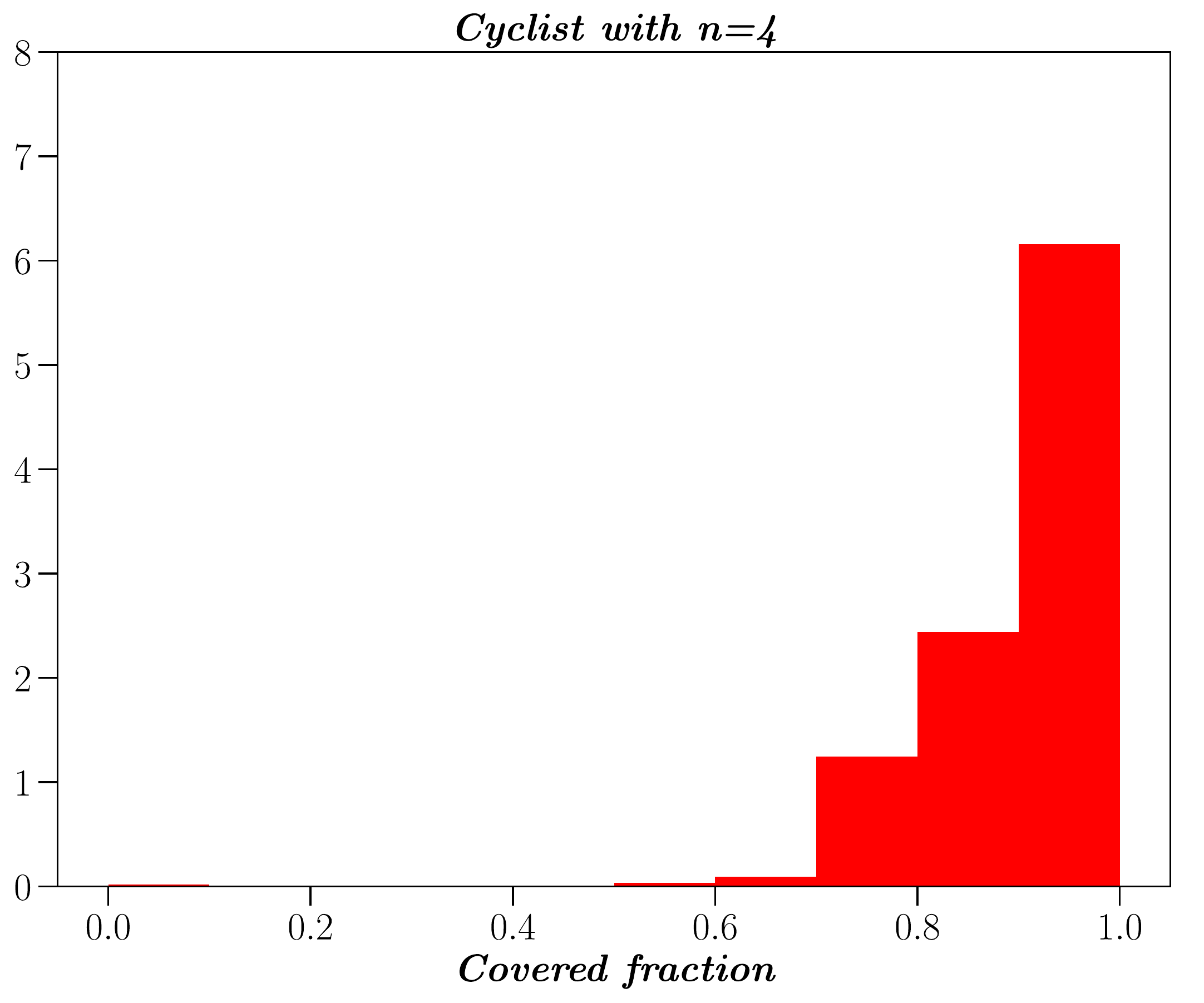}
        \includegraphics[height=1.45in,width=1.45in]{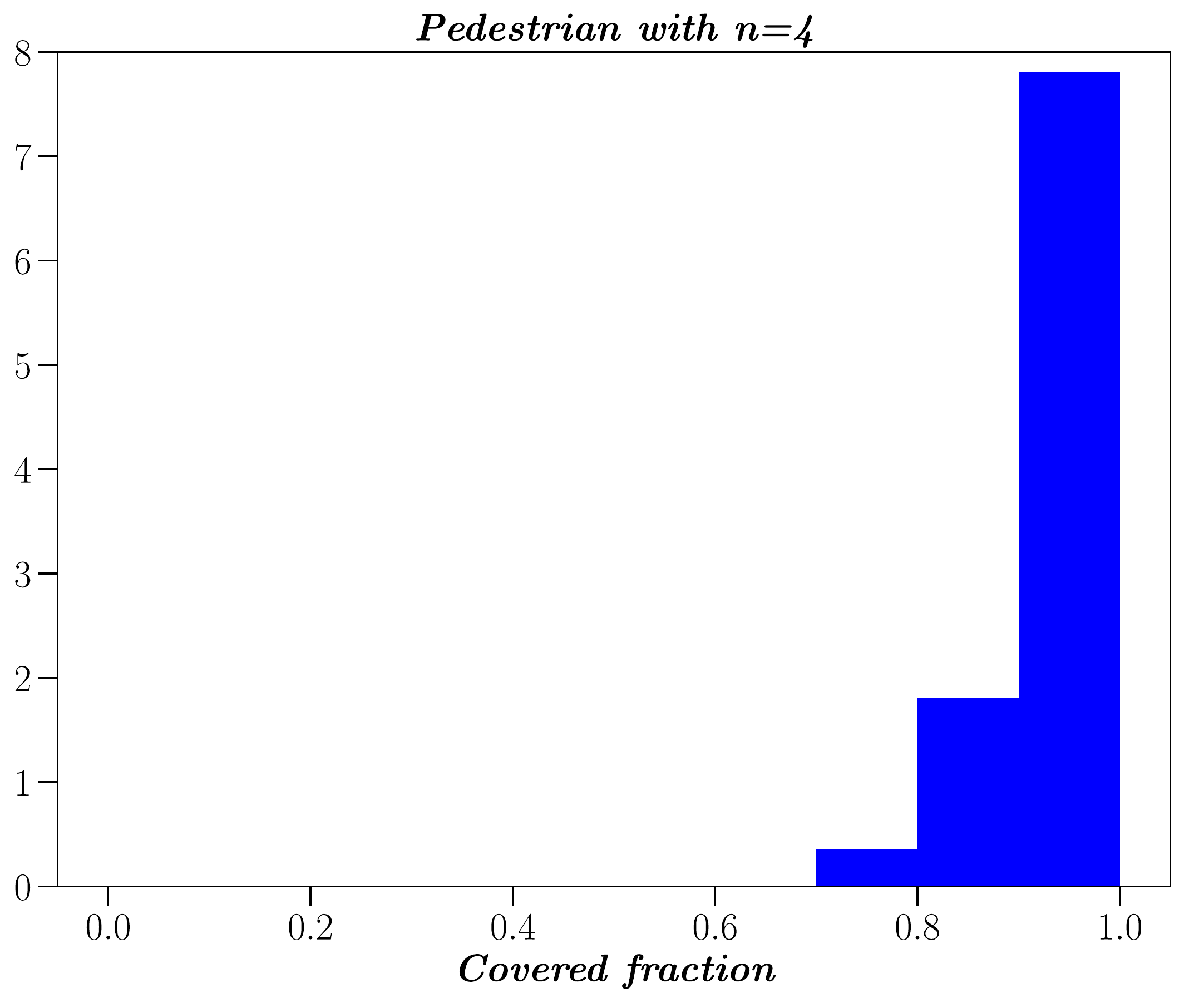}
        \caption{4 Clusters of K-Mean Clustering (n=4)}
    \end{subfigure}
    \begin{subfigure}[t]{.5\textwidth}
        \centering
        \includegraphics[height=1.45in,width=1.45in]{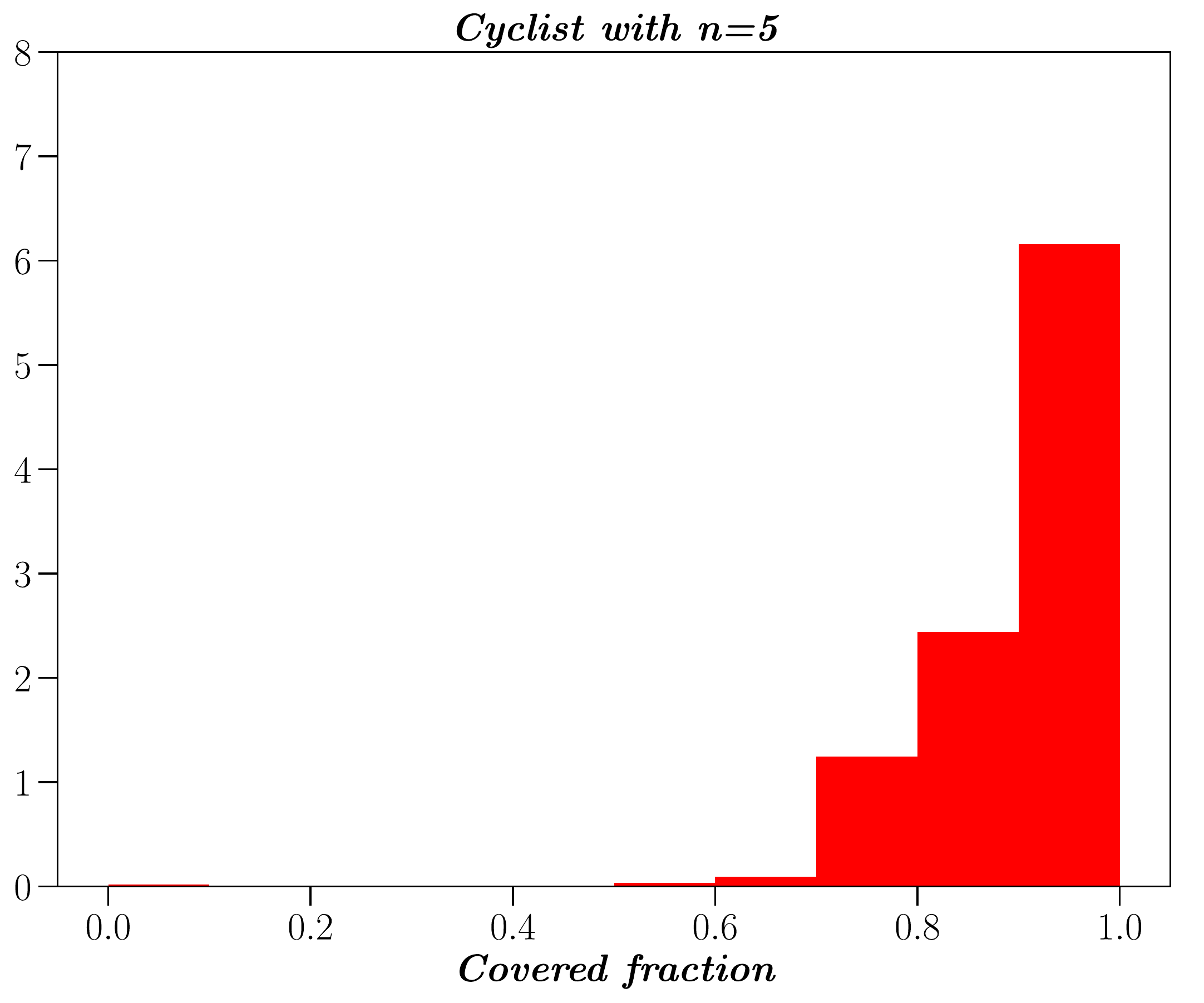}
        \includegraphics[height=1.45in,width=1.45in]{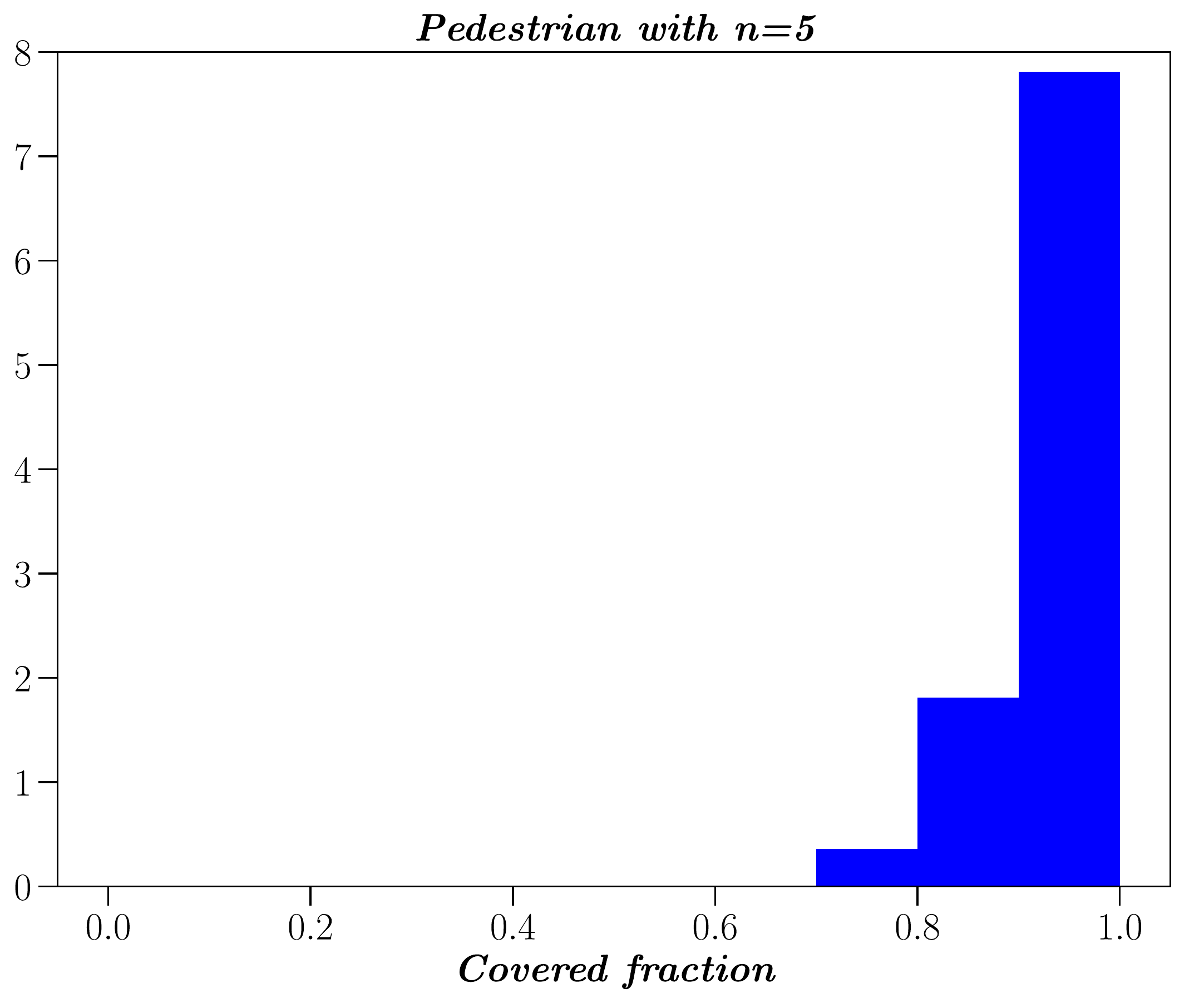}
        \caption{5 Clusters of K-Mean Clustering (n=5)}
    \end{subfigure}
	\caption{Histogram distribution of overlapped area between ground truth box and anchor proposal for Cyclist and Pedestrian objects for different number of K-Mean Clusters as (a) $n=1$, (b) $n=2$, (c) $n=3$, (d) $n=4$ and (e) $n=5$ where $n$ is the number of clusters/groups for $K$-mean clustering method. The x-axis represent the percentage overlapped area between ground truth box and anchor proposal.}
	\label{frac_ground_truth}
\end{figure}

In order to evaluate the anchoring effect we have computed the overlapped area of ground truth and anchors proposal by the proposed CAP in the validation set for various number of cluster $n$ used in $K$-mean clustering method. Figure \ref{frac_ground_truth} illustrates the normalized distribution of the overlapped area of ground truth and proposed anchors for different numbers of clusters in the classes of cyclist and pedestrian. In this Figure, we calculated the overlapped area of ground truth objects in the validation set according to Equation \ref{eq:1} and plotted its histogram distribution for each class. As it can be noted, the distribution in Figure \ref{frac_ground_truth} (a) shows that the proposed anchors for $n=1$ mostly overlapped $50\%$ to $65\%$ of the ground truth objects which makes it more difficult for RPN and final detection network to capture the objects. However, this trend changes significantly as we increase the number of groups ($n$) in $K$-mean clustering algorithm. Figure \ref{frac_ground_truth} (d) and (e) show the overlapped area distribution for $n=4$ and $n=5$, respectively. The results show that significant amount of proposed anchors and objects are overlapped above $85\%$ with high number of clusters during anchoring process that reduces the probability of the missed objects by the model. This effect can also be observed on recall number of RPN and Average Precision of the entire model in next sections. 

\subsection{Region Proposal Network Evaluation} \label{RPNevaluation}

\begin{figure*}[h]
    \centering
    \begin{subfigure}[t]{.99\textwidth}
        \centering
        \includegraphics[height=1.in,width=3.2in]{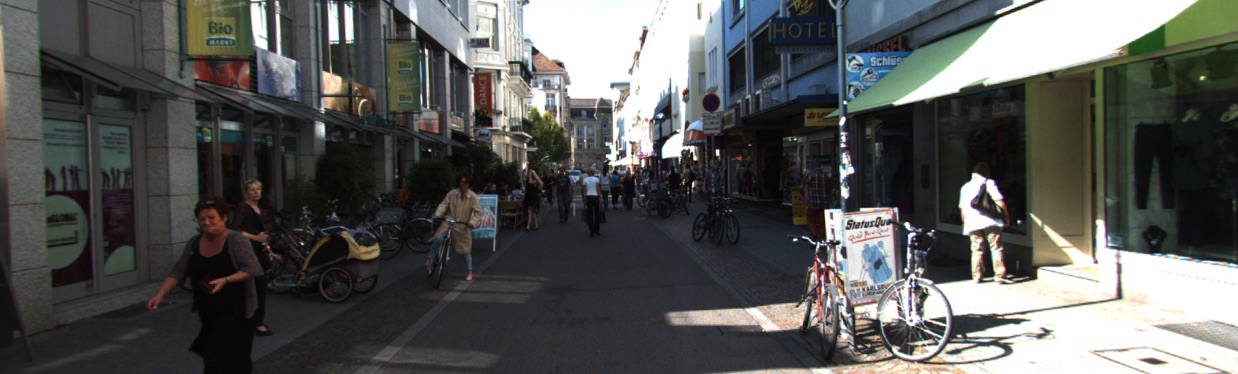}
        \includegraphics[height=1.in,width=3.2in]{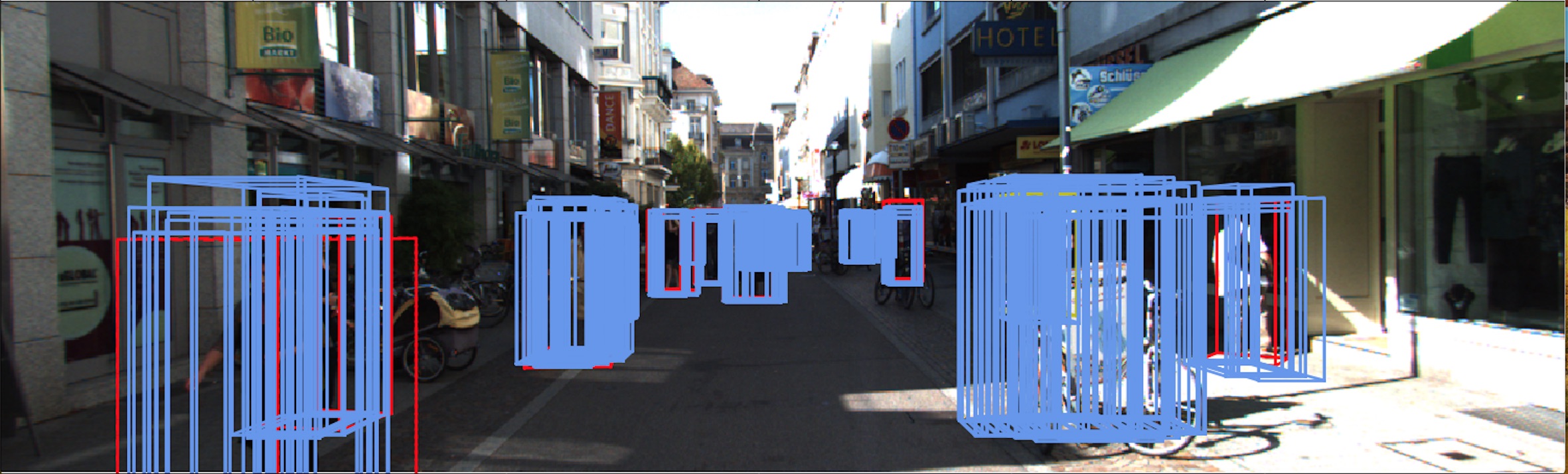}
    \end{subfigure}
    \begin{subfigure}[t]{.99\textwidth}
        \centering
        \includegraphics[height=1.in,width=3.2in]{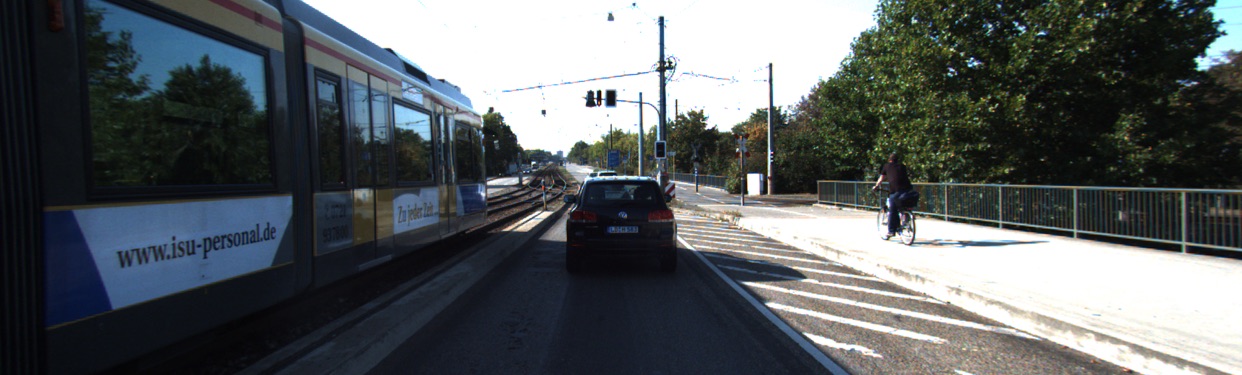}
        \includegraphics[height=1.in,width=3.2in]{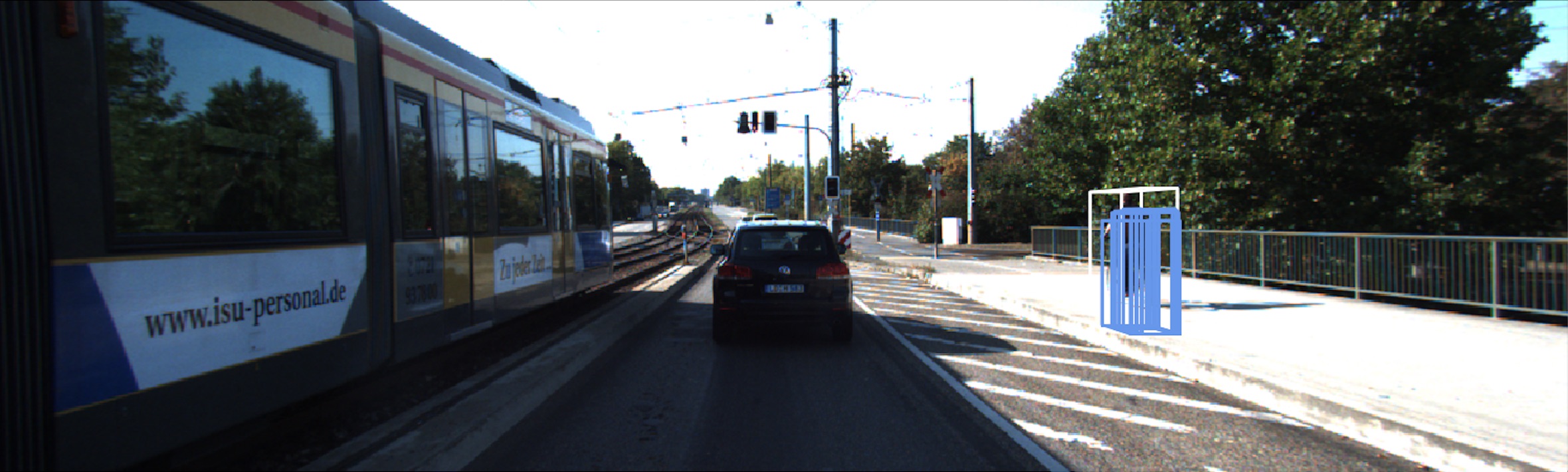}
    \end{subfigure}
    \caption{3D Proposed Region of Interests (ROIs) from RPN. Left column contains images of cyclists and pedestrian objects in RGB images and right column contains images which have the proposed ROIs from RPN of CAP-AVOD. }
	\label{roi}
\end{figure*}

As it was mentioned before, the RPN modifies and adjusts the proposed anchors to bring the attention of the last network toward the areas where the desired objects are located. Hence, for the object which is missed in the CAP the RPN fails to propose the ROI that contains that object and consequently the last network also fails to detect it. Therefore, it is extremely important that the ROIs proposed by the RPN capture the potential areas where the desired objects are located which allows the last network to modify and offset the ROIs easier and detect the objects accurately. Figure \ref{roi} shows 3D proposed ROIs by the RPN for cyclist and pedestrians in the image. In this regard, the performance of the RPN is evaluated by computing the recall according to the proposed ROIs and the ground truth objects. 

\begin{figure}[h]
    % \begin{subfigure}{.5\textwidth}
        \centering
        \includegraphics[height=1.6in,width=1.6in]{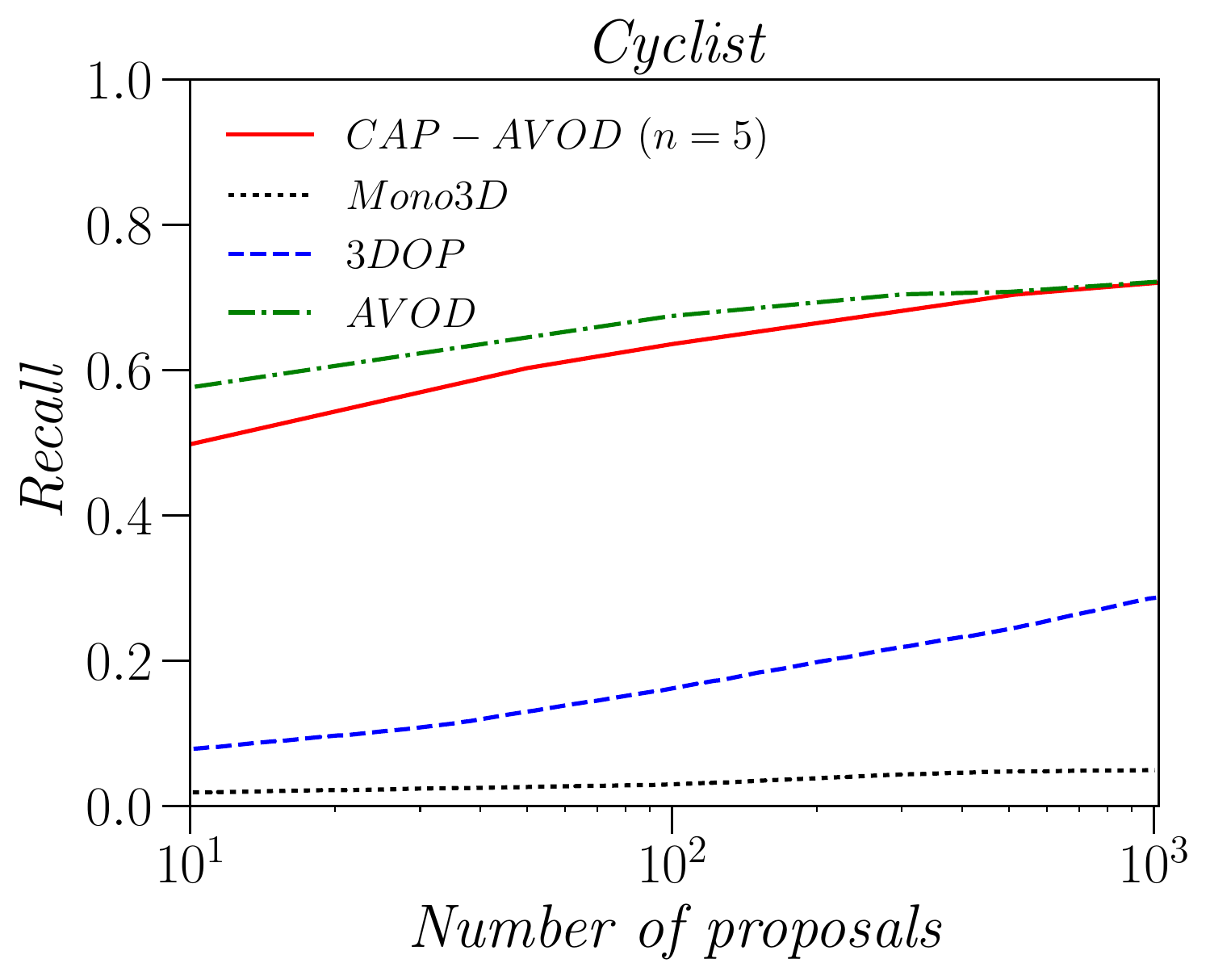}
        \includegraphics[height=1.6in,width=1.6in]{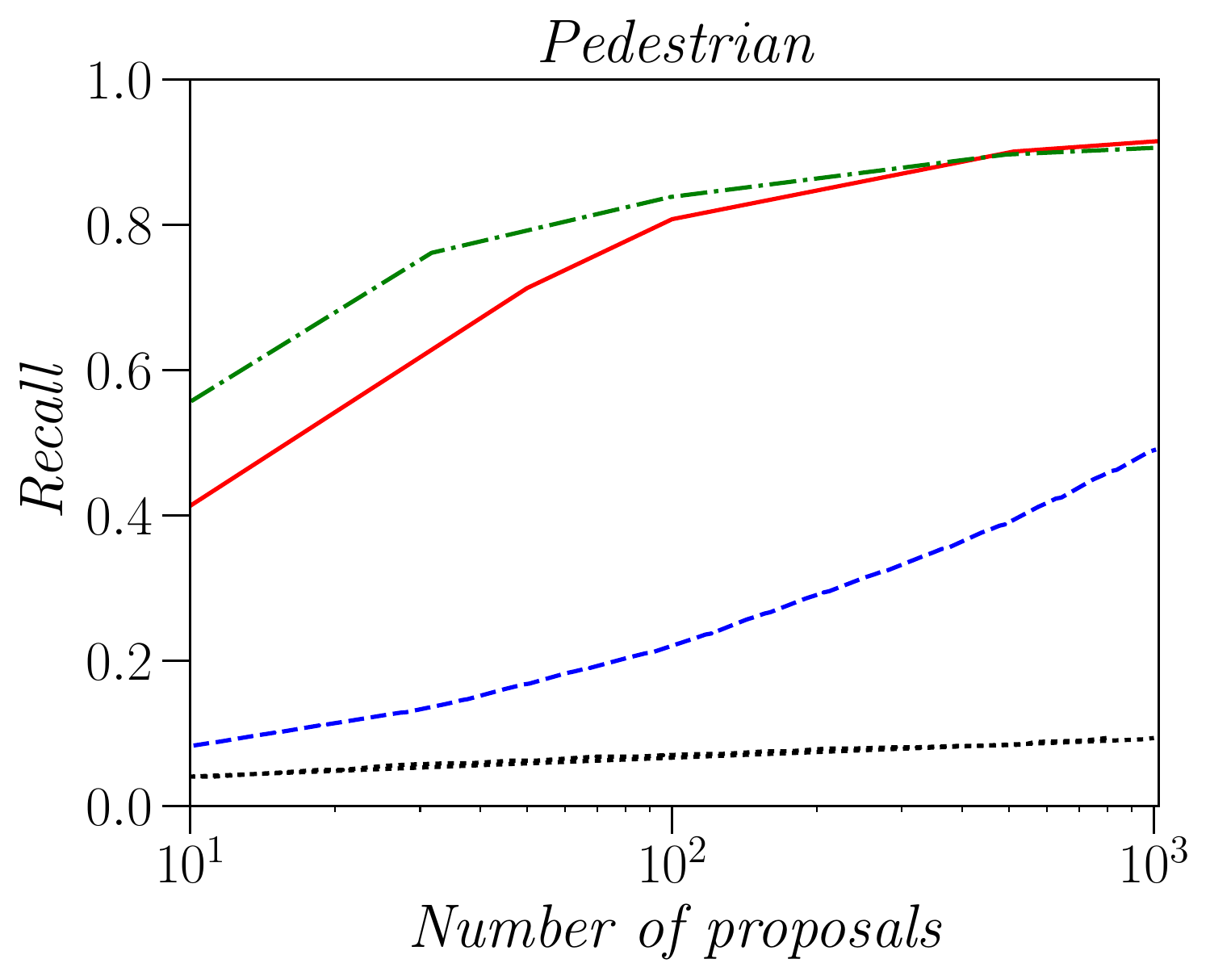}
        ~
        \includegraphics[height=1.6in,width=1.6in]{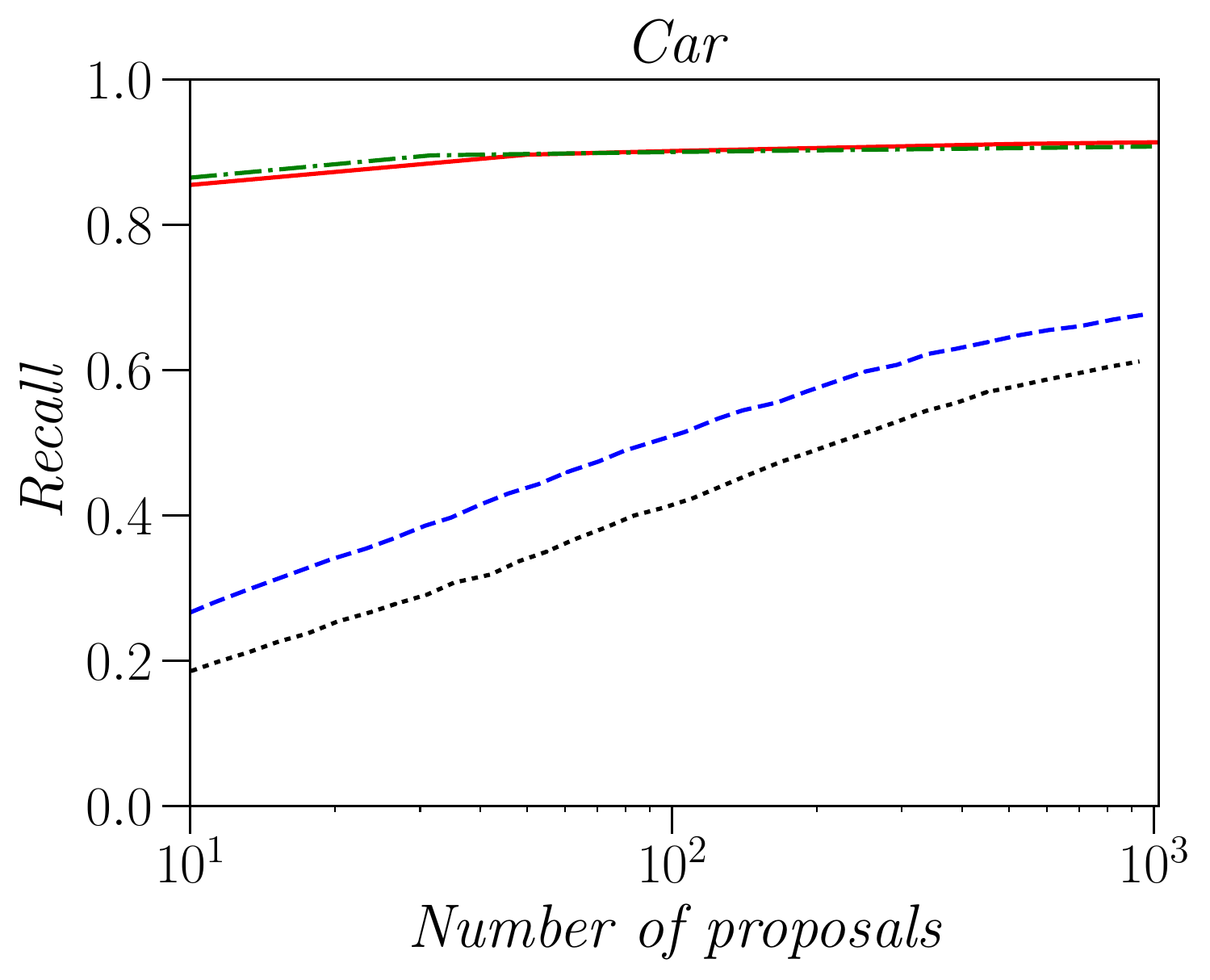}
    \caption{Recall versus number of ROI proposals for Cyclist,  Pedestrian and car computed using state-of-the-art models: Mono3D, 3DOP and AVOD) and CAP-AVOD.}
	\label{Recall}
\end{figure}

Figure \ref{Recall} shows recall as function of ROI proposals number by RPN for Mono3D \cite{chen2016monocular}, 3DOP \cite{chen20153d}, AVOD and CAP-AVOD framwork. For computing recall of CAP-AVOD, we set the number of clusters to $5$ ($n=5$) in anchoring process. The recall is computed based on $0.5$ 3D Intersection-Over-Union (IoU) threshold for classes of cyclist, pedestrian and car.   

According to this figure, there is a significant gap between recall numbers of CAP-AVOD and Mono3D and 3DOP. For instance, recall for Cyclist class reaches to $49.8\%$ at $10$ proposals of CAP-AVOD, while the recall for $3DOP$ is $29\%$ at $1000$ proposals. Similar trend exists for the class of Pedestrian in which the maximum recall that $3DOP$ achieves at $1000$ proposals can be gained with AVOD with $10$ proposals. Furthermore, the RPN in CAP-AVOD reaches to the recall of $90.5\%$ at 1024 proposals for pedestrian which shows the significant enhancement in proposing ROIs in CAP-AVOD. The significant improvement of the recall for the class of the car is also observed in this figure where the recall reaches $91.3\%$ at $1024$ ROIs. Comparison between AVOD and CAP-AVOD indicates that at low number of ROI proposals still AVOD has higher recall for class of cyclist and pedestrians, while with $1024$ ROI proposals the recall of the CAP-AVOD is higher than the AVOD by $~2\%$ for cyclist and have the same value for pedestrian and car classes.           

\subsection{Object Detection Evaluation} \label{detectionevaluation}

As it has been shown in above sections, increasing the number of clusters in CAP strategy, improved the overlapped area of proposed regions with the ground truth objects and recall of the RPN significantly. In this section we report the result of 3D object detection with CAP-AVOD and compare with other state-of-the-art methods (F-PointNet\cite{qi2017pointnet++}, VoxelNet\cite{zhou2017voxelnet} and AVOD\cite{ku2017joint}). Table \ref{table_2d_3d} shows the Average Precision (AP) of 3D object detection for pedestrian, cyclist and car for anchor proposing using K-mean clustering method.
\begin{table}[h]
    \centering
    \caption{3D Average Precision (AP) for Cyclist and Pedestrian and Car at 3D IoU of $0.5$}
    \scalebox{1}{
        \fontsize{6pt}{12pt}
        \selectfont\def\arraystretch{1.5}{
            \setlength{\tabcolsep}{0.5em}
            \begin{tabular}{ |c||c|c|c|c|c|c|c|c|c| }
                \hline
                & \multicolumn{3}{|c|}{Cyclist} & \multicolumn{3}{|c|}{Pedestrian}&
                \multicolumn{3}{|c|}{Car}\\
                \hline
                &Easy&Moderate&Hard&Easy&Moderate&Hard&Easy&Moderate&Hard\\
                \hline
                VoxelNet&61.22&48.36&44.37&39.48&33.69&31.51&77.47&65.11&57.73\\
                \hline
                F-PointNet&71.96&\textbf{56.77}&\textbf{50.39}&51.21&44.89&40.23&81.20&70.39&62.19 \\
                \hline
                AVOD&64.00&52.18&46.61&50.80&42.81&40.88&81.94&71.88&66.38\\
                \hline
                $n=1$&60.10&37.48&33.94&47.55&43.93&41.46&76.81&67.167&66.09\\
                \hline
                $n=2$&68.83&42.40&41.08&53.45&48.25&43.66&\textbf{84.13}&\textbf{74.05}&\textbf{67.75}\\
                \hline
                $n=3$&67.35&44.05&40.31&53.70&47.60&42.67&83.27&74.10&67.96\\
                \hline
                $n=4$&67.88&45.77&40.90&56.93&50.19&43.63&83.89&73.22&67.29\\
                \hline
                $n=5$&\textbf{76.10}&49.01&42.57&\textbf{57.99}&\textbf{50.94}&\textbf{48.08}&82.36&72.99&66.897\\
                \hline
            \end{tabular}
        }
    }
    \label{table_2d_3d}
\end{table}
The result shows that increasing the number of clusters increases the final AP of the model for classes of cyclist, pedestrian and car significantly. This enhancement is more significant for the classes of objects that have more variation in the size such as pedestrian and cyclist. For instance, we achieved average $16.1\%$, $11.54\%$ and $8.63\%$ enhancement on AP respectively on Easy, Moderate and Hard setting of these two classes. We also compare against the AVOD result provided directly by evaluation server. The results shows that increasing the number of clusters in anchoring strategy improves the detection such that for $n=5$ the trained model outperforms the AVOD by $7.19\%$, $8.13\%$ and $8.8\%$ on Easy, Moderate and Hard setting of the pedestrian class. It also has enhancement in the detection by $12.1\%$ increase in AP on Easy setting of cyclist class. The results suggest that the increase in the average precision is more remarkable for the classes of cyclist and pedestrian compared to that of car. This behavior is attributed to the large variation in the size of the objects in cyclist and pedestrian classes which requires anchors to have various geometries that provides the model with appropriate prior for unusual object size in these classes. On the other hand, the size and the aspect ratio of the objects in car class are similar and the model does not need anchors with different size for detecting objects. Hence, the average precision of the model keeps improving with increasing the number of the clusters proposed for anchoring for pedestrian and cyclist classes, however the model's performance saturates at small number of clusters for the car class.  
\begin{table}[h]
    \centering
    \caption{3D Average Precision (AP) for Cyclist and Pedestrian and Car at 3D IoU of $0.5$ for anchor proposing with GMM}
    \scalebox{1}{
        \fontsize{6pt}{12pt}
        \selectfont\def\arraystretch{1.5}{
            \setlength{\tabcolsep}{0.5em}
            \begin{tabular}{ |c||c|c|c|c|c|c|c|c|c| }
                \hline
                & \multicolumn{3}{|c|}{Cyclist} & \multicolumn{3}{|c|}{Pedestrian}&
                \multicolumn{3}{|c|}{Car}\\
                \hline
                &Easy&Moderate&Hard&Easy&Moderate&Hard&Easy&Moderate&Hard\\
                \hline
                $n=1$&50.18&32.82&32.27&48.08&46.84&42.21&77.34&67.31&65.97\\
                \hline
                $n=2$&66.67&48.77&39.96&53.51&46.96&43.09&77.41&72.78&66.85\\
                \hline
                $n=3$&68.57&42.54&41.69&55.96&48.92&44.11&83.16&73.57&67.35\\
                \hline
                $n=4$&73.61&48.51&46.39&57.22&50.49&46.31&83.61&73.90&67.42\\
                \hline
                $n=5$&68.62&43.00&41.33&57.74&50.85&46.14&82.62&72.95&66.74\\
                \hline
            \end{tabular}
        }
    }
    \label{table_2d_3d-GMM}
\end{table}

Furthermore, 3D Average Precision for the models that use anchoring with GMM is also reported in Table \ref{table_2d_3d-GMM}. The results have similar trend as it is observed for models that use K-mean clustering in anchoring process. As it is noted, increasing the number of clusters leads to significant increase in the final average precision.
\begin{figure*}[t]
    \centering
    \begin{subfigure}[t]{.99\textwidth}
        \centering
        \includegraphics[height=1.in,width=3.2in]{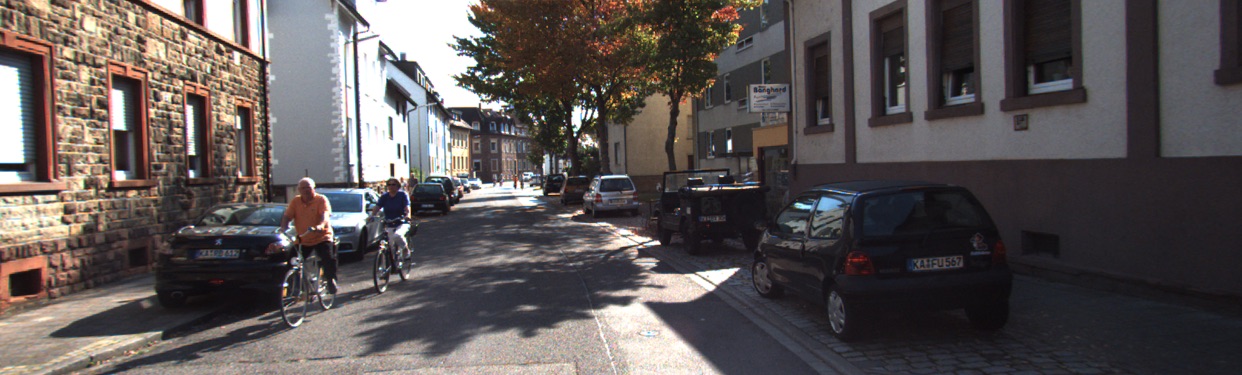}
        \includegraphics[height=1.in,width=3.2in]{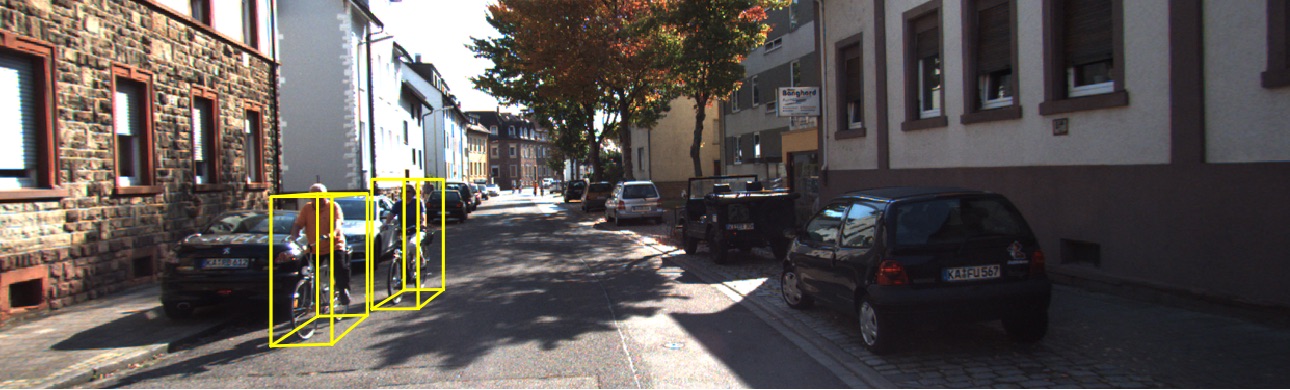}
    \end{subfigure}
    \begin{subfigure}[t]{.99\textwidth}
        \centering
        \includegraphics[height=1.in,width=3.2in]{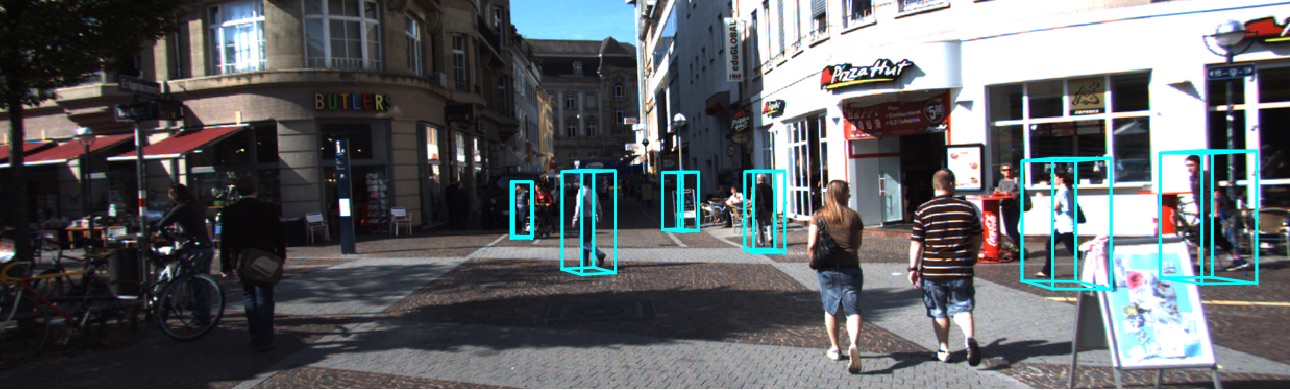}
        \includegraphics[height=1.in,width=3.2in]{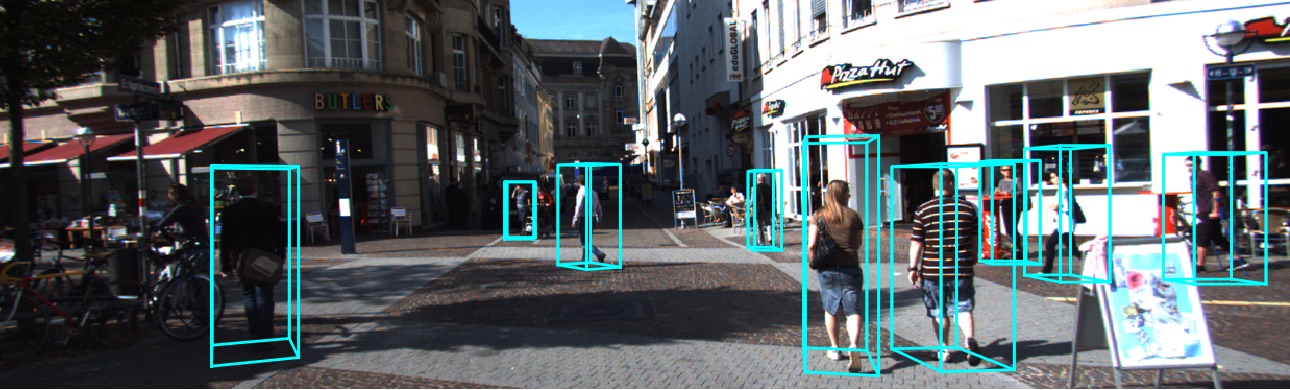}
        \end{subfigure}
    \begin{subfigure}[t]{.99\textwidth}
        \centering
        \includegraphics[height=1.in,width=3.2in]{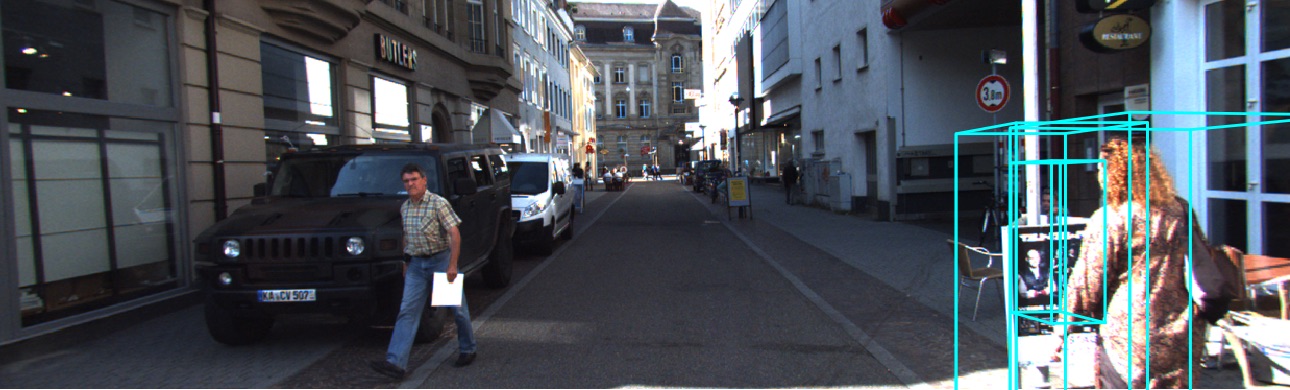}
        \includegraphics[height=1.in,width=3.2in]{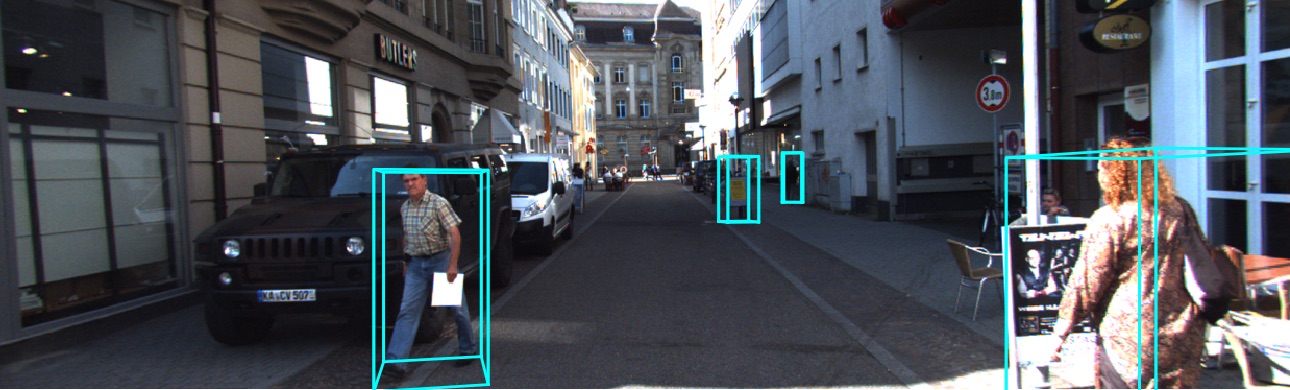}
    \end{subfigure}
    \caption{Result of 3D Object Detection using RGB images and LIDAR Point Cloud with AVOD and CAP-AVOD frameworks. Left column contains images which show 3D detected object with AVOD framework \cite{ku2017joint}. Right column contains images which show 3D detected objects with CAP-AVOD framework. }
	\label{detection}
\end{figure*}
The prior size of the anchors is set by the mean values of data points falling into each cluster determined by K-mean or GMM clustering methods. Hence, it is expected that the mean dimension for the each cluster is far from the dimensions of the objects with abnormal size and aspect ratio. Therefore, these objects are missed in the final detection. This behavior is attributed to the presence of the objects with abnormal sizes in cyclist class which are not similar to the majority of objects in this class. It explains the drop in the AP on moderate and hard setting of cyclist class for $n=5$ compared to that of AVOD model. Furthermore, if we want to have anchors with similar size to those with unusual dimensions we should set larger number of clusters in clustering method which increases the number of the anchors significantly leading to higher computation. Furthermore, we show the performance of the model on the class of car. As it is noted, the AP increases by $2.19\%$, $2.17\%$ and $1.27\%$ on Easy, Moderate and Hard settings of car class, respectively. This maximum AP occurs for $n=2$ that is caused by the fact that the size distribution of the objects in car class can be clustered better with $n=2$ compared to other number of clusters. Figure \ref{detection} shows the effect of anchoring on the final detection. According to this figure the detection of the pedestrians and cyclist has been improved significantly. 
Hence, our results suggest that $n=2$ is an appropriate cluster number for proposing anchors for the class of the cars, while $n=5$ is the best cluster number for the classes of cyclist and pedestrian as it enables the model to boost its performance significantly.

%===========================================================
\section{Conclusions and Future Work}
%===========================================================

In this paper, we propose class specific anchoring proposal strategy to improve the 3D object detection with AVOD framework. The proposed anchoring strategy used K-Mean and GMM clustering for each class of object which increased the AP of 3D object detection in combined RGB and point cloud data set (KITTI) by $7.19\%$, $8.13\%$ and $8.8\%$ on Easy, Moderate and Hard setting of the pedestrian class and $12.1\%$ on Easy setting of cyclist class. We also illustrate that the clustering in anchoring process also enhances the performance of the RPN in proposing ROIs significantly. We conclude that $n=2$ is an appropriate cluster number in proposing anchors for the class of the cars, while $n=5$ is the best cluster number for the classes of cyclist and pedestrian as it enables the model to boost its performance significantly. In future work, we intend to study the performance of single-stage object detectors and compare them to multi-stage object detectors with the hope to further improve the AP.

\bibliographystyle{aaai}
\bibliography{egbib.bib}

\end{document}